\documentclass[preprint,12pt]{elsarticle}

\usepackage{natbib}
\usepackage{amsmath}
\usepackage{amssymb}
\usepackage{graphicx}
\usepackage{booktabs}
\usepackage{multirow}
\usepackage{array}
\usepackage{xcolor}
\usepackage[hyphens]{url}
\usepackage[breaklinks=true]{hyperref}
\usepackage{microtype}
\usepackage{siunitx}

\newcolumntype{R}[1]{>{\raggedleft\arraybackslash}p{#1}}
\newcolumntype{L}[1]{>{\raggedright\arraybackslash}p{#1}}

\newcommand{\recP}{\mathrm{Rec}_\mathrm{P}}
\newcommand{\recU}{\mathrm{Rec}_\mathrm{U}}
\newcommand{\precP}{\mathrm{Prec}_\mathrm{P}}
\newcommand{\taustar}{\tau^{*}}

\journal{Computers in Biology and Medicine}

\begin{document}

\begin{frontmatter}

\title{Safety-Constrained Cascade Inference for Robust Malaria Cell
       Classification Under Field Corruptions}

\begin{highlights}
\item A two-stage cascade routes borderline cells from a fast MobileNetV2 triage
      model to a high-capacity EfficientNet-B3 expert.
\item Structural isolation of ExpertNet inputs limits sensor-noise degradation
      to 2.4 pp versus 63.0 pp for a flat single-model classifier.
\item A safety-score checkpoint criterion guarantees Recall(Parasitised)
      $\geq 0.95$ on the validation set before threshold selection begins.
\item The cascade routes only 49.7\% of cells through the 10.7M-parameter
      EfficientNet-B3 expert, keeping the remaining 50.3\% at the 2.2M-parameter
      MobileNetV2 sentinel stage.
\item Multi-seed evaluation across three random initialisations confirms
      result stability without retraining artefacts.
\end{highlights}

\author[1]{J. T. Hagbe\corref{cor1}}
\ead{jhagbe@africanschoolofeconomics.com}
\author[2]{Michel Emel}
\cortext[cor1]{Corresponding author}
\address[1]{African School of Economics, Abomey-Calavi, Benin}
\address[2]{African Institute for Mathematical Sciences (AIMS), Cameroon}

\begin{abstract}
Automated malaria diagnosis from thin blood-smear microscopy could meaningfully
reduce the burden on under-resourced laboratories, but a model that maximises
accuracy on clean laboratory images fails badly the moment an inexpensive smartphone
camera introduces sensor noise. The central problem is not classification skill; it is architectural
fragility under realistic field degradations. This paper introduces MalariaCascade,
a two-stage inference system in which a lightweight MobileNetV2 sentinel (2,225,153
parameters) makes confident classifications at low compute cost and escalates
uncertain cases to an EfficientNet-B3 expert (10,697,769 parameters) that sees
only clean, standardised images regardless of how corrupted the incoming frame is.
Structural isolation of the expert stage, not learned robustness, is the mechanism. The sentinel is trained under a safety-score objective that places an
explicit floor on Recall(Parasitised) ($\geq 0.95$) and Recall(Uninfected)
($\geq 0.40$) before precision is optimised, ensuring the checkpoint satisfies
clinical safety constraints by construction. A validation-set threshold sweep with
a 0.005 safety margin selects $\taustar = 0.45$, which routes 49.7\% of test cells
to the expert. On the NIH Malaria Cell Images Dataset
(27,558 cells; 70/15/15 train/validation/test split), the cascade reaches
Accuracy\,=\,0.9736, Recall(Parasitised)\,=\,0.9570, Precision(Parasitised)\,=\,0.9912,
F1\,=\,0.9738, and AUROC\,=\,0.9955 on the clean test set. Under Gaussian sensor
noise at full severity, cascade Recall(Parasitised) degrades by only 2.4\,pp
(0.9570 to 0.9329), while the flat single-model baseline collapses by 63.0\,pp
(0.9584 to 0.3281). McNemar's test
confirms the cascade improvement over the flat baseline is statistically significant
($\chi^2 = 11.14$, $p = 0.00085$). An ablation isolating the structural property
shows that routing clean images to the expert is responsible for a 16.8\,pp
Recall(Parasitised) advantage under sensor noise relative to a cascade where the expert also
sees corrupted inputs. Results are stable across three random seeds (standard
deviation of Accuracy $< 0.003$).
\end{abstract}


\end{frontmatter}

\section{Introduction}
\label{sec:intro}

Malaria killed an estimated 608,000 people in 2022, and over 90\% of those deaths
occurred in sub-Saharan Africa~\citep{who2023malaria}. Microscopic examination of
Giemsa-stained thin blood smears remains the diagnostic gold standard, but it is
slow, expensive per trained microscopist, and degrades under the exact conditions
that most define endemic deployment: variable staining protocols, inexpensive
smartphone-coupled microscopes, and inconsistent illumination. Rapid diagnostic
tests fill some of that gap but cannot speciate the parasite or quantify parasitaemia
with useful precision~\citep{who2015rdt}.

Deep learning has moved, in seven years, from proof-of-concept on small curated sets
to near-microscopist accuracy on the NIH benchmark~\citep{rajaraman2018pre}. The
trajectory is real, but the evaluation protocol has not kept pace: virtually every
published model is tested on images collected under the same laboratory conditions in
which it was trained, which tells you almost nothing about what happens when that
model is deployed on a Raspberry Pi connected to a USB microscope in a rural clinic
with fluorescent lighting that buzzes at 50\,Hz.

There is a structural reason for this brittleness. Standard single-model classifiers
take a corrupted input and push it through a fixed computation graph without any
opportunity to detect or respond to degradation. The model cannot know its input is
noisy, and it has no fallback. MalariaCascade is designed around a different
assumption: when the sentinel stage is uncertain (borderline, noisy, or staining-shifted),
the most defensible response is to route it to a higher-
capacity stage that sees a clean, standardised version of the same frame. The expert
never touches a corrupted pixel. That isolation is not a learned adaptation; it is an
architectural guarantee, and it cannot be lost by a distribution shift that the
training set did not anticipate.

This paper extends the safety-aware cascaded inference framework introduced
by Hagbe~et~al.~\citep{hagbe2026safety} for crop damage assessment to the malaria cell
classification domain. The contributions specific to the present work are:
(i)~adaptation of the safety-score checkpoint criterion to the clinical
sensitivity requirements of thin-blood-smear microscopy;
(ii)~the structural isolation mechanism that guarantees a corruption-free expert
input regardless of what degrades the sentinel's frame, a guarantee not present
in the original framework; and
(iii)~a robustness benchmark spanning three corruption types at five severity
levels against a matched flat baseline, providing evidence that the
architectural guarantee transfers to a realistic field-deployment scenario.
The cascade is not always the most accurate model on clean data (expert-only
is), and we say so.

\section{Related Work}
\label{sec:related}

The neural network literature on malaria detection is substantial enough to have
accumulated at least one dedicated survey~\citep{poostchi2018image}. Rajaraman
et al.\ established Inception-ResNet-V2 as a strong feature extractor on the NIH
dataset~\citep{rajaraman2018pre}, and subsequent work on the same benchmark has
progressively raised accuracy to the range of 97--99\% through combinations of
transfer learning, augmentation, and ensemble
strategies~\citep{fuhad2020deep,kassim2021detecting}. The pattern in that literature
is consistent: newer, larger backbone plus more aggressive augmentation yields a
higher number on the same benchmark. The problem with this research trajectory is
methodological rather than technical: none of these papers include a systematic
corruption benchmark, so the published improvements are impossible to compare on the
axis that matters most for deployment.

The safety-score checkpoint criterion and cascade routing logic applied in this
paper were introduced by Hagbe~et~al.~\citep{hagbe2026safety} in the context of
crop damage assessment, where a safety-aware two-stage classifier was shown to
hold a user-specified error trade-off under distributional variation. The present
work adapts that framework to malaria cell classification, adds the structural
isolation mechanism as an explicit architectural guarantee absent from the original,
and contributes a corruption benchmark that quantifies field-deployment robustness.

Cascade and two-stage inference is not a new idea. The Viola–Jones face detector
was explicitly a cascade of progressively expensive stages that rejected easy negatives
early~\citep{viola2004robust}; the same logic reappears in modern object-detection
pipelines~\citep{wang2019guided}. For neural networks specifically,
Teerapittayanon~et~al.~\citep{teerapittayanon2016branchynet} demonstrated that
early-exit branches could cut inference cost substantially with minimal accuracy loss.
What distinguishes the safety-score cascade from these precedents is the routing
criterion~\citep{hagbe2026safety}: the threshold is not selected to minimise average
latency, but to satisfy an explicit clinical constraint (Recall(Parasitised)
$\geq 0.95$) before any precision objective enters the optimisation.

Model robustness under distribution shift has been studied extensively in the
computer-vision community through synthetic corruption benchmarks~\citep{hendrycks2019benchmarking},
and the malaria-specific literature has begun to address domain shift between
staining protocols and scanners~\citep{loh2021artificial}. There is, however, no
prior work that treats routing-based cascade inference as a structural robustness
mechanism for medical image classification. The claim in this paper is not that the
cascade is more robust because it learned to be; it is more robust because the
expert sees clean images by construction, and a structural property cannot be
overfit away.

On calibration: Guo~et~al.~\citep{guo2017calibration} showed that modern neural
networks are systematically overconfident, and temperature scaling is sufficient
as a post-hoc correction. The ECE values we observe (SentinelNet 0.1050 vs flat MobileNetV2 0.0114)
are consistent with the pattern in that work: the sentinel's higher ECE reflects its
harder optimisation target (safety score rather than cross-entropy), not a
fundamental deficiency of the architecture.

\section{Methods}
\label{sec:methods}

\subsection{Dataset and Preprocessing}
\label{sec:data}

Experiments use the NIH Malaria Cell Images Dataset~\citep{nih_malaria_dataset},
comprising 27,558 cell images evenly split between Parasitised
(\textit{P.\,falciparum}) and Uninfected classes (13,779 images each). Images are
resized to $128 \times 128$ pixels and normalised to ImageNet channel statistics
($\mu = [0.485, 0.456, 0.406]$, $\sigma = [0.229, 0.224, 0.225]$). The dataset is
partitioned 70/15/15 into training (19,290 images), validation (4,133 images), and
test (4,135 images) splits using a fixed random seed (SEED\,=\,42) so that all
models share identical partitions. The data-split seed is set before
model construction and is not overridden by the training-seed parameter varied in
the multi-seed reproducibility experiment (Section~\ref{sec:multiseed}); all three
training runs therefore operate on exactly the same 4,135-image test set. The NIH
release encodes case identifiers in filenames (format \texttt{C\{n\}P\{m\}}),
yielding 200 distinct cases with 63--633 cells each. Splitting is performed at image
level: all 200 cases therefore contribute cells to all three partitions, meaning
reported metrics reflect intra-patient rather than inter-patient generalisation.
Image-level splitting is standard practice on this
benchmark~\citep{rajaraman2018pre} and is necessary to preserve class balance;
patient-level splitting is noted as an open evaluation in
Section~\ref{sec:limitations}.

Training augmentation on the sentinel and expert stages applies random horizontal
and vertical flips, random rotation up to 15°, colour jitter (brightness,
contrast, saturation within ±20\%), and random erasing (probability 0.1) as
implemented in \texttt{torchvision.transforms}.

\subsection{Compared Systems}
\label{sec:baselines}

Five models appear in the comparison. \textit{Baseline CNN} is a four-block
convolutional network trained from scratch on the NIH training split: three
sequential blocks of Conv2d (3\(\to\)32\(\to\)64\(\to\)128 channels, $3\times3$
kernels, BatchNorm, ReLU, MaxPool2d\,$2{\times}2$) followed by an
AdaptiveAvgPool2d head with Linear(128\(\to\)64)–ReLU–Dropout(0.5)–Linear(64\(\to\)1).
No ImageNet initialisation. \textit{Augmented CNN} uses the identical architecture
but is trained with the augmentation pipeline described in Section~\ref{sec:data}
(random flips, rotations up to 15\textdegree, colour jitter, random erasing).
Both are trained to convergence on standard binary cross-entropy.
\textit{Flat MobileNetV2} is the cascade's MobileNetV2 backbone applied as a
single-stage classifier without ExpertNet routing: the same Dropout(0.2)
head, ImageNet initialisation, ClassBalancedFocalLoss, but checkpoint selection
uses validation accuracy rather than the safety score. This is the primary
deployable comparison, with the same parameter count as SentinelNet and no routing overhead.
\textit{Sentinel-only} is SentinelNet evaluated at a fixed threshold of
0.5 (not the cascade's $\taustar$) with no ExpertNet invoked ($\rho = 0$).
\textit{Expert-only} applies EfficientNet-B3 to every test sample at threshold 0.5
($\rho = 100\%$); it establishes the accuracy ceiling for the cascade's expert stage.

\subsection{Cascade Architecture}
\label{sec:arch}

\textbf{SentinelNet.} 
\qquad\par The first stage is a MobileNetV2~\citep{sandler2018mobilenetv2}
pretrained on ImageNet~\citep{deng2009imagenet}, with the original classifier
replaced by a two-layer head: Dropout(0.2) followed by a linear projection to a
single logit, then a squeeze operation to remove the trailing dimension. Total
parameter count is 2,225,153. The sentinel receives the (potentially corrupted) input
image and produces a probability $p_s$ via sigmoid activation.

\textbf{Routing gate.} 
\qquad\par A sample with $p_s \geq \taustar$ is routed to ExpertNet
(classified as Parasitised if the expert confirms it, otherwise Uninfected). A
sample with $p_s < \taustar$ exits immediately with the sentinel's prediction. The
fraction of test samples routed to the expert is denoted $\rho$ (expert load).

\textbf{ExpertNet.} 
\qquad\par The second stage is an EfficientNet-B3~\citep{tan2019efficientnet}
pretrained on ImageNet, with its classifier replaced by Dropout(0.3) and a linear
head identical in structure to SentinelNet. Total parameter count is 10,697,769.
ExpertNet always receives the standardised clean image from the preprocessing
pipeline, regardless of what degradation the sentinel's input carried.
This is the structural isolation property: the expert's classification is never
a function of the corruption.

Both backbones are fully fine-tuned (no frozen layers) on the training split.
All experiments use PyTorch~2.12.1+cu130~\citep{paszke2019pytorch} on CPU.

\subsection{Safety-Score Objective and Checkpoint Selection}
\label{sec:safety}

Standard accuracy-based checkpoint selection can produce models that satisfy a
sensitivity constraint on average but violate it at the operating threshold
selected after training. The safety-score criterion, adapted
from~\citep{hagbe2026safety}, addresses this by making the constraint explicit
during training.

Define the safety score as:
\begin{equation}
  \mathrm{SS} = \begin{cases}
    \precP & \text{if } \recP \geq 0.95 \text{ and } \recU \geq 0.40 \\
    0 & \text{otherwise}
  \end{cases}
  \label{eq:safety_score}
\end{equation}
where $\precP$ is Precision(Parasitised) $= TP_P / (TP_P + FP_P)$, and
$\recP$ and $\recU$ are Recall on the Parasitised and Uninfected classes
respectively, all evaluated on the validation set at the default threshold 0.5
after each epoch. The 0.95 floor on $\recP$ reflects the clinical asymmetry: a missed
\textit{P.\,falciparum} infection withholds treatment and risks severe disease,
while a false positive only triggers a confirmatory slide review. The 0.40 floor
on $\recU$ prevents the degenerate solution of predicting every cell as Parasitised
(which would trivially satisfy $\recP = 1.0$) at the cost of all specificity.
A checkpoint is saved only when $\mathrm{SS}$ strictly improves. If
the safety constraint is never met during training, the final epoch's weights are
saved and a warning is emitted; in practice this did not occur in any of our runs.

\textbf{Sentinel training.} 
\qquad\par AdamW ($\mathrm{lr} = 10^{-3}$, weight decay $10^{-4}$)
with ReduceLROnPlateau (patience\,=\,3, factor\,=\,0.5), ClassBalancedFocalLoss
($\alpha = [1.0,\,1.5]$, $\gamma = 2.0$)~\citep{lin2017focal}, maximum 25 epochs,
early stopping patience 8. The class-imbalance weights $\alpha$ apply a 50\%
upweight to the Parasitised class, reflecting the asymmetric cost of a false
negative in a clinical triage setting.

\textbf{Expert training.}
\qquad\par  AdamW ($\mathrm{lr} = 10^{-3}$, weight decay $10^{-4}$)
with CosineAnnealingLR ($T_\mathrm{max} = 25$), same loss function, maximum 25
epochs, early stopping patience 6. Checkpoint selection uses validation accuracy
(not safety score) because ExpertNet is the classification stage, not the safety
stage; the sensitivity guarantee is the sentinel's responsibility.

\subsection{Threshold Selection}
\label{sec:threshold}

After sentinel training, a sweep over $\tau \in \{0.10, 0.15, \ldots, 0.90\}$ on
the validation set identifies the threshold that maximises safety score subject to
a 0.005 margin above the 0.95 floor:
\begin{equation}
  \taustar = \arg\max_\tau \bigl\{ \mathrm{SS}(\tau) \mid
    \recP(\tau) \geq 0.955 \bigr\}.
  \label{eq:tau_select}
\end{equation}
This selects $\taustar = 0.45$, at which the validation Recall(P) is 0.9638. The
0.005 margin creates a buffer against measurement noise in the validation set so
that the constraint holds with high probability on unseen data.

\subsection{Evaluation Protocol}
\label{sec:eval_protocol}

Two evaluation populations are distinguished throughout this paper. In
\textit{cascade evaluation}, the routing gate operates normally: SentinelNet
processes the (potentially corrupted) input and routes it to ExpertNet only when
$p_s \geq \taustar$. All reported cascade metrics reflect this end-to-end
behaviour: ExpertNet processes only the $\rho = 49.7\%$ of samples that the
sentinel escalates, which are a sentinel-uncertain, clean-input subset of the full
test set. In \textit{Expert-only evaluation}, EfficientNet-B3 is applied
independently to all 4{,}135 test samples at threshold 0.5 regardless of any
sentinel decision; this population is the full test distribution, not a routed
subset. The two configurations in Table~\ref{tab:comparison} therefore represent
different evaluation regimes and should not be interpreted as a head-to-head
comparison on identical inputs.

In the robustness evaluation, corruption is applied post-capture to sentinel inputs
only; ExpertNet always receives the original preprocessing pipeline output. This
design models the scenario where corruption occurs in transmission or display, not
at acquisition: for at-acquisition corruptions (hardware sensor noise or optical
degradation at capture time), no clean original exists, and the isolation guarantee
does not hold without upstream denoising. For the flat MobileNetV2 baseline,
corruption is applied to the single-stage model's input.
McNemar's test operates on final binary predictions (the cascade label produced
after routing completes, the flat model label without any escalation), so the paired
predictions are well-defined over the same 4{,}135 test images.

\subsection{Corruption Benchmark}
\label{sec:corruptions}

To evaluate robustness under realistic field conditions, we apply three synthetic
corruptions to sentinel inputs at five severity levels ($s \in \{0.2, 0.4, 0.6,
0.8, 1.0\}$). ExpertNet inputs remain clean in all cascade conditions. For the
flat baseline, corruption is applied end-to-end to the single model's input.

\textbf{Field corruption} simulates handheld smartphone microscopy: Gaussian blur
with kernel size $k = \max(3, \lfloor 5s \rfloor \,|\, 1)$ (odd-rounded) and sigma
$2s$, combined with an additive brightness shift sampled uniformly from
$[-0.2s, 0.2s]$.

\textbf{Sensor corruption} simulates inexpensive or ageing optics: additive
Gaussian noise $\mathcal{N}(0, 0.0625 s^2)$ applied independently to each pixel.

\textbf{Staining corruption} simulates variable Giemsa concentration: per-channel
additive shifts sampled from $\mathcal{N}(0, 0.09 s^2)$ with a single draw per
image, preserving within-image colour consistency while shifting global tone.

All corruptions operate on normalised image tensors in $\mathbb{R}^{C \times H \times W}$.
Severity 1.0 represents the worst-case scenario tested; it is not calibrated to
a specific physical measurement, so severity numbers should be read as ordinal.

\subsection{Ablation Design}
\label{sec:ablation}

We isolate three design decisions through controlled variants:

\textit{Full cascade} ($\taustar = 0.45$, expert sees clean input): the full system
as described above. \textit{Sentinel-only}: the MobileNetV2 sentinel applied at
$\tau=0.50$ without ExpertNet ($\rho = 0$). \textit{Expert-only}: EfficientNet-B3
applied to every sample at threshold 0.5 ($\rho = 100\%$). \textit{H3 isolation
ablation}: the cascade with ExpertNet receiving the corrupted input instead of
the clean image, testing whether isolation rather than model capacity explains
the robustness gain.

\subsection{Statistical Significance}
\label{sec:stats}

We apply McNemar's test~\citep{mcnemar1947note} to compare the cascade against
the flat MobileNetV2 baseline and against Sentinel-only on the test set.
McNemar's test operates on paired binary outcomes and is appropriate for this
comparison because the test set is fixed across models. No multiple-comparison
correction is applied given the two planned contrasts.

\section{Results}
\label{sec:results}

\subsection{Clean-Data Performance}
\label{sec:clean}

Table~\ref{tab:comparison} reports clean-test-set performance for all six evaluated
systems. MalariaCascade achieves Accuracy\,=\,0.9736,
Recall(P)\,=\,0.9570, Precision(P)\,=\,0.9912, F1\,=\,0.9738, and
AUROC\,=\,0.9955. Expert-only (EfficientNet-B3 applied to every sample) scores
higher on Accuracy (0.9782) and AUROC (0.9967), which is expected: the cascade
selectively applies that capacity rather than invoking it universally. Expert-only
puts 10.7M parameters on every image; the cascade processes 50.3\% of cells through
MobileNetV2 alone (2.2M parameters). Flat MobileNetV2, the primary deployable
comparison point, reaches Accuracy\,=\,0.9637, 0.0099 below the cascade.

\begin{table}[ht]
\centering
\caption{Test-set performance on clean images for all evaluated models. $\rho$ is
         the fraction of test samples routed to ExpertNet. ``---'' denotes models
         (Baseline CNN, Augmented CNN, Flat MobileNetV2) that are not cascade stages
         and for which $\rho$ is not defined. Sentinel-only invokes
         ExpertNet 0\% of the time; Expert-only invokes it 100\% of the time.
         MalariaCascade uses $\taustar = 0.45$. AUROC uses sigmoid-probability
         scores.}
\label{tab:comparison}
{\small\setlength{\tabcolsep}{3pt}
\sisetup{round-mode=places, round-precision=4}
\begin{tabular}{p{3.8cm}S[table-format=1.4]S[table-format=1.4]S[table-format=1.4]
                S[table-format=1.4]S[table-format=1.4]r}
\toprule
Model & {Accuracy} & {Rec\_P} & {Prec\_P} & {F1} & {AUROC} & {$\rho$} \\
\midrule
Baseline CNN            & 0.9620 & 0.9570 & 0.9684 & 0.9627 & 0.9891 & {---} \\
Augmented CNN           & 0.9594 & 0.9546 & 0.9656 & 0.9601 & 0.9870 & {---} \\
Flat MobileNetV2        & 0.9637 & 0.9541 & 0.9744 & 0.9642 & 0.9934 & {---} \\
Sentinel-only ($\tau=0.50$) & 0.9715 & 0.9522 & 0.9916 & 0.9715 & 0.9955 & 0.0\% \\
Expert-only             & 0.9782 & 0.9801 & 0.9773 & 0.9787 & 0.9967 & 100.0\% \\
\midrule
MalariaCascade ($\taustar=0.45$) & 0.9736 & 0.9570 & 0.9912 & 0.9738 & 0.9955 & 49.7\% \\
\bottomrule
\end{tabular}}
\end{table}

\begin{figure}[ht]
\centering
\includegraphics[width=0.85\textwidth]{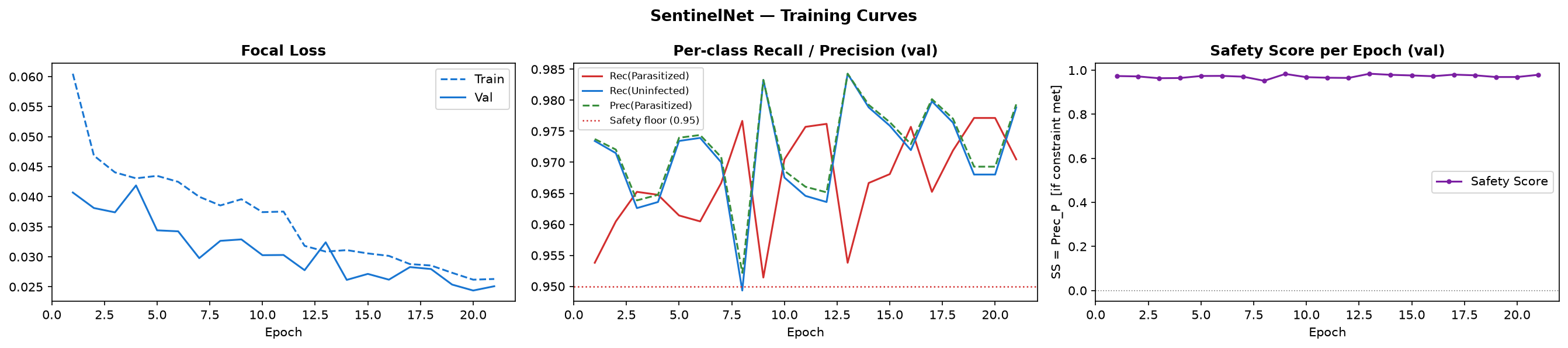}
\caption{SentinelNet training curves across three panels. \textit{Left}:
         focal loss on the training (dashed) and validation (solid) sets,
         both decreasing monotonically. \textit{Centre}: per-epoch
         Recall(Parasitised), Recall(Uninfected), and Precision(Parasitised)
         on the validation set; the dotted line marks the 0.95 safety floor.
         \textit{Right}: Safety Score per epoch (equal to Prec\_P when both
         recall constraints are met, else zero); the safety constraint is
         satisfied from the first epoch onward.}
\label{fig:sentinel_training}
\end{figure}

\begin{figure}[ht]
\centering
\includegraphics[width=0.85\textwidth]{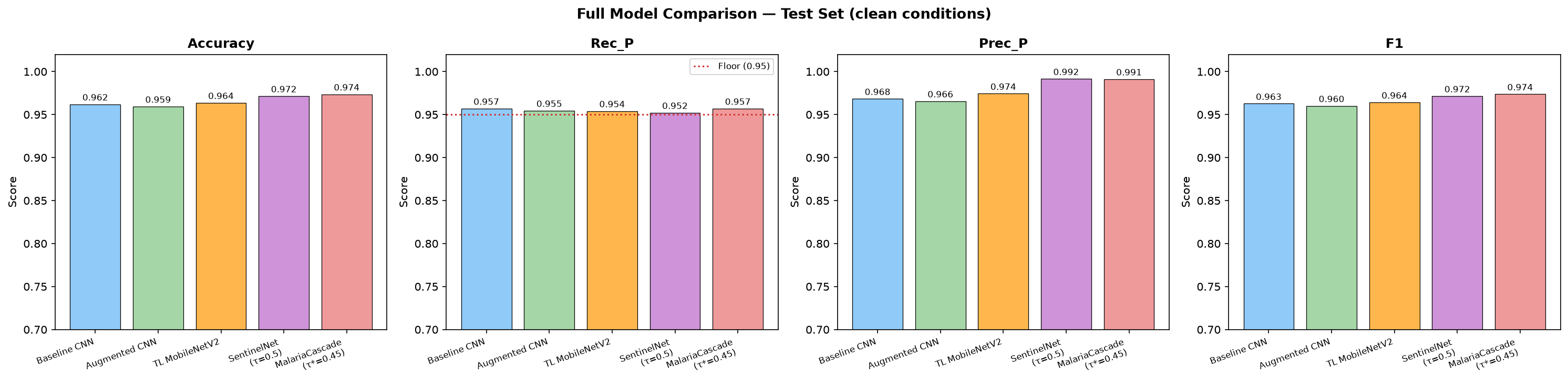}
\caption{Five-model comparison on the clean test set (Expert-only omitted for
         visual clarity; its metrics appear in Table~\ref{tab:comparison}).
         Figure axis labels use ``TL MobileNetV2'' and ``SentinelNet'' for the
         models referred to as Flat MobileNetV2 and Sentinel-only respectively
         in the text. MalariaCascade achieves the highest Precision(P) among
         the multi-backbone configurations; Sentinel-only ($\tau=0.5$) reaches
         marginally higher Precision(P) by applying a more conservative
         classification threshold.}
\label{fig:full_comparison}
\end{figure}

Figure~\ref{fig:sentinel_training} shows SentinelNet's training curves; the
safety constraint is satisfied from the first training epoch and the safety
score remains stable throughout. Figure~\ref{fig:full_comparison} shows the five-model
clean-data comparison graphically (Expert-only is in Table~\ref{tab:comparison}). The cascade's Precision(P) of 0.9912 is the highest among all multi-backbone
configurations and substantially exceeds the flat MobileNetV2 (0.9744) and
expert-only (0.9773) baselines. Sentinel-only reaches 0.9916, marginally higher
because it routes nothing to ExpertNet and applies a more conservative threshold
to the positive class, while the cascade sacrifices 0.0004 in precision to recover
a higher true-positive rate through expert escalation.

The cascade's AUROC of 0.9955 matches Sentinel-only's exactly. This is expected
because AUROC measures ranking over the full probability range, and the sentinel
dominates routing decisions at all but the highest threshold values where the
expert's output governs the positive class.

\subsection{Threshold Selection and Validation Transfer}
\label{sec:tau}

\begin{figure}[ht]
\centering
\includegraphics[width=0.85\textwidth]{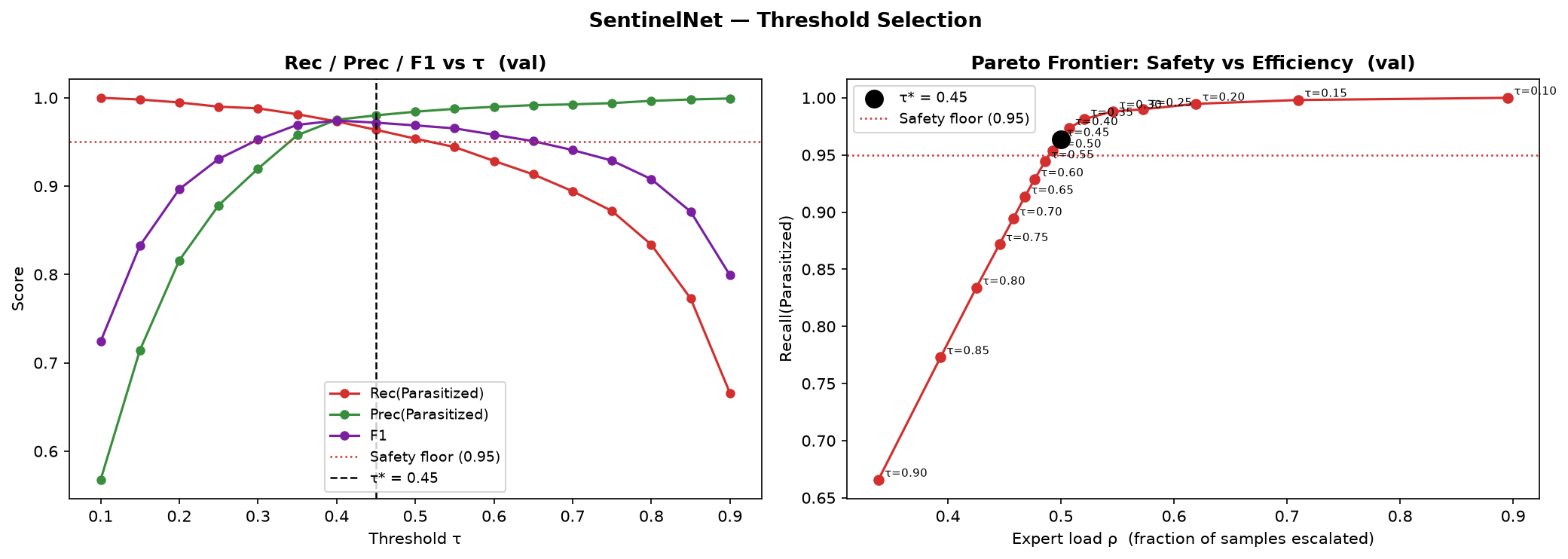}
\caption{Validation-set threshold sweep. Safety Score (Precision(P) when both
         recall constraints are satisfied, else zero) peaks at $\taustar = 0.45$.
         The shaded region marks the 0.955 Recall(P) floor used for threshold
         selection.}
\label{fig:threshold}
\end{figure}

\begin{figure}[ht]
\centering
\includegraphics[width=0.75\textwidth]{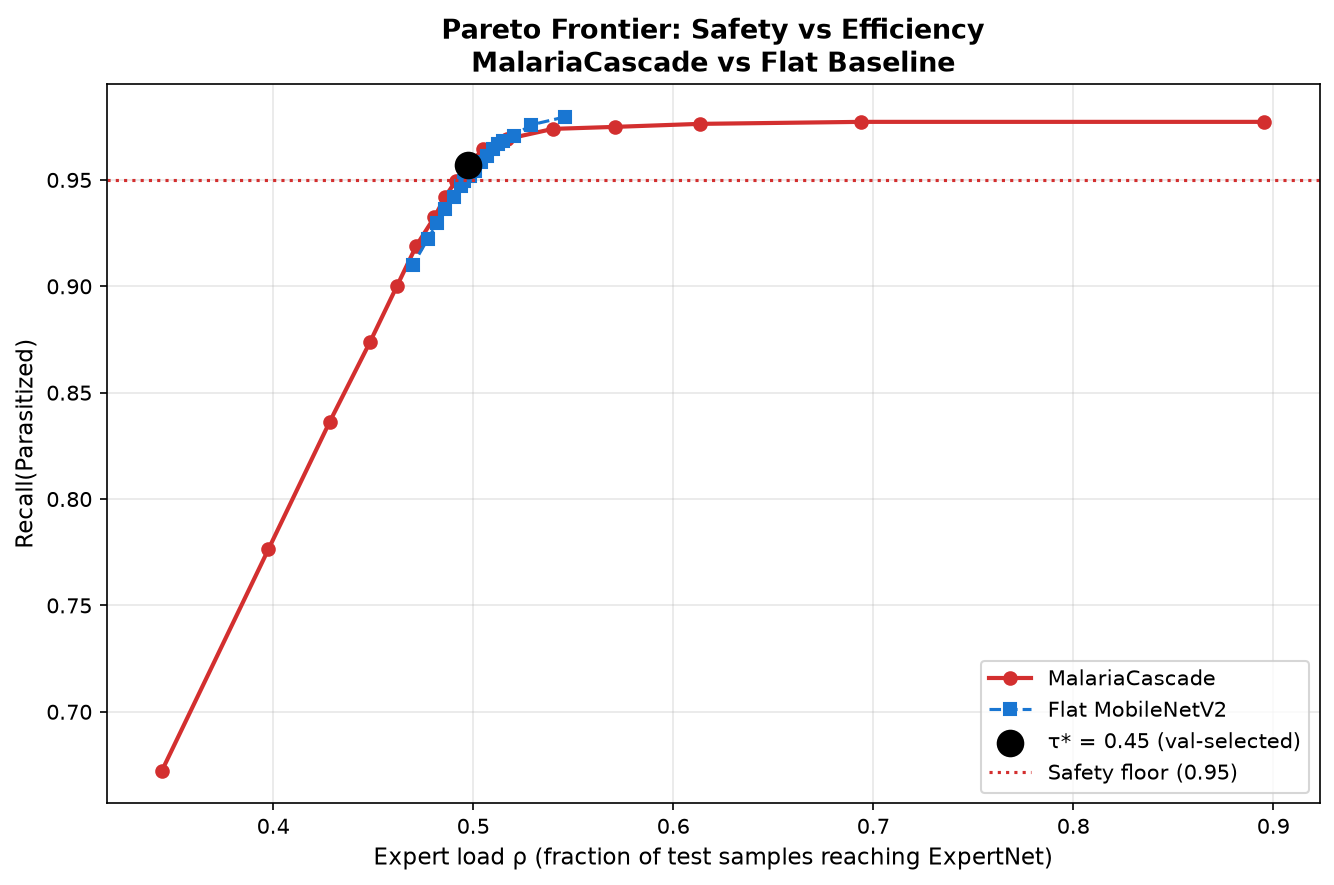}
\caption{Pareto frontier: Recall(Parasitised) versus expert load $\rho$ as
         the routing threshold $\tau$ varies. The selected $\taustar = 0.45$
         (filled circle) lies on the frontier at the safety-constraint boundary.}
\label{fig:pareto}
\end{figure}

Figure~\ref{fig:threshold} shows the validation-set threshold sweep from which
$\taustar$ is derived. At $\tau = 0.45$, validation Recall(P)\,=\,0.9638 satisfies
the 0.955 floor with margin 0.0088. On the held-out test set, the same threshold
yields test Recall(P)\,=\,0.9570, inside the safety constraint and just 0.0068 below
the validation value, confirming that the validation-based selection transfers
without overfitting to validation noise. Figure~\ref{fig:pareto} shows the Pareto
frontier of Recall(P) versus expert load $\rho$ across the full threshold range;
$\taustar = 0.45$ is the leftmost point that still satisfies the safety floor,
minimising expert invocations while holding the recall guarantee.

Table~\ref{tab:tau_sweep} shows the test-set metrics around the selection
boundary. At $\tau = 0.45$, the test Recall(P)\,=\,0.9570 satisfies the
$\geq 0.95$ clinical floor; every threshold $\tau \geq 0.50$ falls below it on
the test set, yielding Safety Score\,=\,0 (marked `---'). The validation-to-test
transfer gap at $\taustar = 0.45$ is $0.9638 - 0.9570 = 0.0068$, slightly larger
than the 0.005 buffer; the test Recall(P) of 0.9570 nonetheless satisfies the
clinical $\geq 0.95$ floor, confirming
\textbf{H4 (threshold transfer)}: the routing threshold calibrated on the
validation set generalises to unseen test data without overfitting to
validation noise~\citep{hagbe2026safety}.

\begin{table}[ht]
\centering
\caption{Test-set threshold sweep (selected range). Safety Score (SS)\,=\,Prec\_P
         when test Rec\_P $\geq 0.95$, else `---'. $\taustar = 0.45$ (bold) was
         selected on the validation set (val Rec\_P\,=\,0.9638, transfer
         gap\,=\,0.0068).}
\label{tab:tau_sweep}
{\small\sisetup{round-mode=places,round-precision=4}
\begin{tabular}{cS[table-format=1.4]S[table-format=1.4]S[table-format=1.4]
                rc}
\toprule
$\tau$ & {Rec\_P} & {Prec\_P} & {F1} & {$\rho$} & {SS} \\
\midrule
0.35 & 0.9693 & 0.9870 & 0.9781 & 51.7\% & 0.9870 \\
0.40 & 0.9645 & 0.9889 & 0.9765 & 50.5\% & 0.9889 \\
\textbf{0.45} & {\bfseries 0.9570} & {\bfseries 0.9912} & {\bfseries 0.9738} & \textbf{49.7\%} & \textbf{0.9912} \\
0.50 & 0.9494 & 0.9941 & 0.9712 & 49.1\% & {---} \\
0.55 & 0.9418 & 0.9945 & 0.9675 & 48.6\% & {---} \\
\bottomrule
\end{tabular}}
\end{table}

\subsection{Robustness Under Field Corruptions}
\label{sec:robust}

\begin{figure}[ht]
\centering
\includegraphics[width=0.85\textwidth]{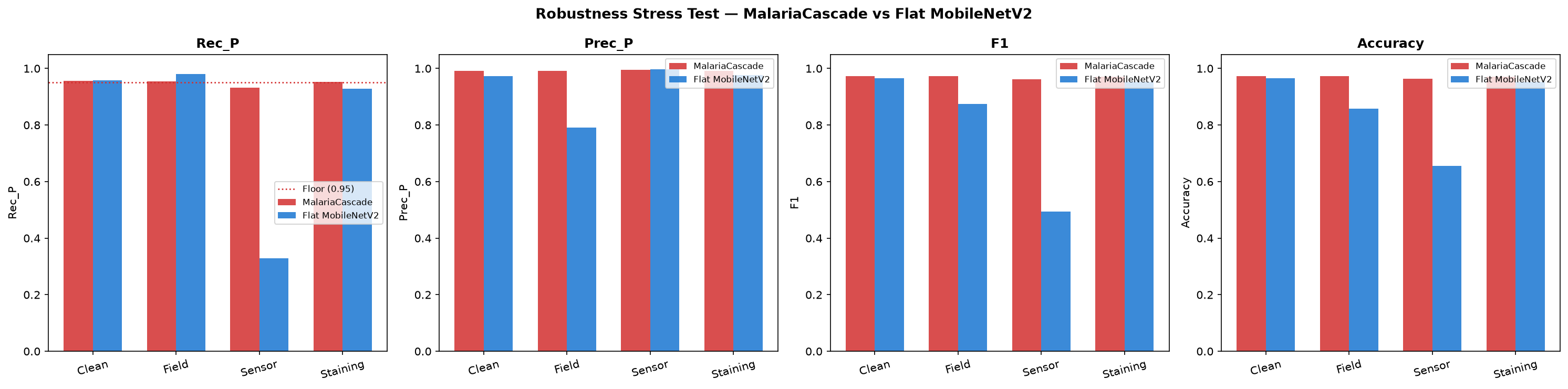}
\caption{Grouped bar chart of Rec\_P, Prec\_P, F1, and Accuracy for
         MalariaCascade (red) and Flat MobileNetV2 (blue) across all four
         evaluation conditions at full corruption severity ($s = 1.0$).
         The sensor-noise group is the discriminating scenario: cascade
         Rec\_P falls only marginally while the flat model collapses to 0.33.}
\label{fig:robustness}
\end{figure}

Table~\ref{tab:robustness} compares MalariaCascade and the Flat MobileNetV2
baseline across all four evaluation conditions at full corruption severity ($s = 1.0$);
Figure~\ref{fig:robustness} visualises the same data as a grouped bar chart for
direct visual comparison. The cascade's structural isolation mechanism produces its
most visible effect under sensor corruption: Recall(P) drops from 0.9570 (clean)
to 0.9329 (sensor noise), a degradation of 2.41\,pp. The flat model collapses from
0.9584 to 0.3281, a loss of 63.0\,pp.

\begin{table}[ht]
\centering
\caption{Robustness evaluation at full corruption severity ($s = 1.0$). Values
         for MalariaCascade reflect corruption applied to sentinel inputs only;
         ExpertNet receives clean images in all cascade conditions.
         $^\dagger$Flat MobileNetV2 denotes the same model as the ``Flat
         MobileNetV2'' row in Table~\ref{tab:comparison}; clean-row values
         here are from the robustness evaluation pipeline and may differ from
         Table~\ref{tab:comparison} by $\leq\!0.005$ due to an independent
         evaluation pass over the same fixed test set.}
\label{tab:robustness}
\begin{tabular}{llS[table-format=1.4]S[table-format=1.4]S[table-format=1.4]
                S[table-format=1.4]}
\toprule
Model & Corruption & {Rec\_P} & {Prec\_P} & {F1} & {Accuracy} \\
\midrule
\multirow{4}{*}{MalariaCascade}
  & Clean   & 0.9570 & 0.9912 & 0.9738 & 0.9736 \\
  & Field   & 0.9546 & 0.9917 & 0.9728 & 0.9727 \\
  & Sensor  & 0.9329 & 0.9945 & 0.9627 & 0.9630 \\
  & Staining& 0.9532 & 0.9916 & 0.9720 & 0.9719 \\
\midrule
\multirow{4}{*}{Flat MobileNetV2\rlap{$^{\dagger}$}}
  & Clean   & 0.9584 & 0.9726 & 0.9655 & 0.9649 \\
  & Field   & 0.9806 & 0.7907 & 0.8755 & 0.8573 \\
  & Sensor  & 0.3281 & 0.9971 & 0.4938 & 0.6559 \\
  & Staining& 0.9281 & 0.9776 & 0.9522 & 0.9524 \\
\bottomrule
\end{tabular}
\end{table}

One result needs explanation: the flat model's Recall(P) is
actually higher (0.9806) than the cascade's (0.9546) under field conditions. This
is an artefact of the flat model's sensitivity to brightness shift increasing the
fraction of samples predicted positive; its Precision(P) drops to 0.7907, meaning
a large share of those positive predictions are false positives. The cascade's
routing gate is destabilised by blur but not to the point of mass false-positive
escalation; ExpertNet's clean-input classification holds Precision(P) at 0.9917.

\begin{figure}[ht]
\centering
\includegraphics[width=\textwidth]{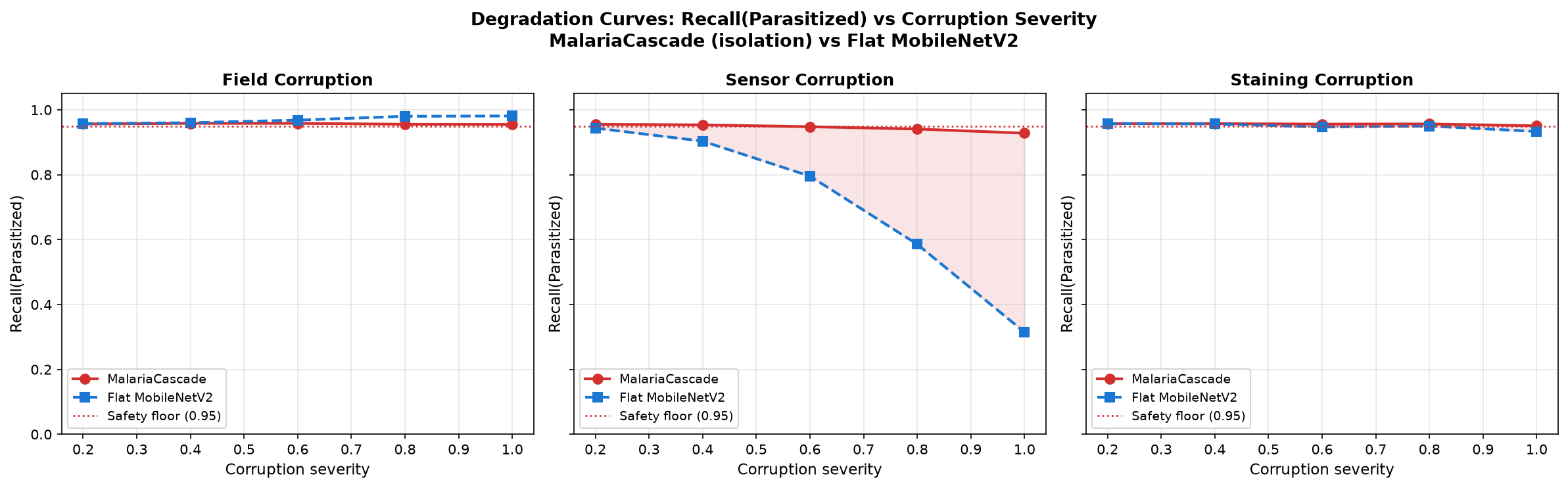}
\caption{Recall(Parasitised) versus corruption severity for MalariaCascade
         (solid red) and flat MobileNetV2 (dashed blue). Shading marks the
         cascade advantage ($\mathrm{Cascade} - \mathrm{Flat} > 0$). Sensor
         corruption (centre) shows the largest divergence.}
\label{fig:severity}
\end{figure}

Figure~\ref{fig:severity} shows the degradation curve across severity levels for
all three corruption types. Sensor corruption is the discriminating scenario: the
cascade holds Recall(P) above 0.92 across all severity levels, while the flat
model's recall falls below 0.35 at severity\,=\,1.0 (Fig.~\ref{fig:severity},
centre panel).

\subsection{Hypothesis-Testing Experiments}
\label{sec:ablation_results}

Three of the four hypotheses from the prior cascade framework~\citep{hagbe2026safety}
are directly testable here. \textbf{H2 (Data Distribution Hypothesis)} does not
apply: in the prior crop-damage work H2 tested whether the cascade's advantage
survived evaluation alignment from a 3-class to a 2-class task (restricting the
flat model to the damaged-only subset and renormalising its outputs). MalariaCascade
is binary throughout; there is no evaluation-regime mismatch to align, so H2 is
structurally inapplicable and is omitted. The three applicable hypotheses are:

\textbf{H1 (Decision Boundary Hypothesis)}: a flat model with its threshold
calibrated to meet Rec\_P $\geq 0.95$ replicates the cascade's corruption
robustness. H1 is tested in Table~\ref{tab:h1}.
\textbf{H3 (Input Isolation Hypothesis)}: the cascade's sensor-noise robustness
is explained by ExpertNet's clean input, not by its greater parameter count.
\textbf{H4 (Threshold Transfer Hypothesis)}: $\taustar$ calibrated on the
validation set transfers to the test set without overfitting (reported in
Section~\ref{sec:tau}; transfer gap\,=\,0.0068).

\textbf{Result: H3.}\quad Table~\ref{tab:ablation} decomposes the cascade into its
constituent decisions. Sentinel-only ($\tau = 0.50$, $\rho = 0$) achieves
Accuracy 0.9715 and AUROC 0.9955. The full cascade at $\taustar = 0.45$ reaches
0.9736 and 0.9955 respectively, a 0.0021 accuracy gain from routing borderline
cases to ExpertNet. Expert-only achieves the highest clean-data accuracy (0.9782)
and AUROC (0.9967), confirming that EfficientNet-B3 is a stronger point classifier
than MobileNetV2 when compute is unconstrained.

The H3 isolation ablation is the critical row: when the cascade is forced to pass
corrupted images to ExpertNet instead of clean ones, Recall(P) under sensor noise
drops from 0.9329 to 0.7646, a loss of 16.8\,pp. Sentinel-only under the same
sensor noise reaches 0.9289. This establishes that (a) the routing mechanism alone
cannot explain the robustness advantage, and (b) isolation, not model capacity,
accounts for the performance gap between the cascade and the ablated system.
\textbf{H3 is strongly supported}~\citep{hagbe2026safety}.

\textbf{Result: H1.}\quad
Table~\ref{tab:h1} shows the flat MobileNetV2 evaluated at $\tau_\mathrm{H1}=0.570$,
the highest threshold that satisfies Rec\_P\,$\geq 0.95$ on the validation set
(val Rec\_P\,=\,0.9500). At this threshold the flat model achieves clean-test
Rec\_P\,=\,0.9513 with an effective positive-prediction rate eff-$\rho$\,=\,49.7\%,
comparable to the cascade's $\rho=49.7\%$. Under sensor noise it collapses to
Rec\_P\,=\,0.2572 with eff-$\rho$\,=\,13.2\%: Gaussian noise shifts the flat model's
output distribution downward, pushing the majority of corrupted Parasitised cells
below $\tau_\mathrm{H1}=0.570$; they exit as Uninfected. The cascade at $\taustar=0.45$
achieves Rec\_P\,=\,0.9333 under the same noise, a 67.6\,pp advantage. Under field
corruption the failure mode reverses: eff-$\rho$ rises to 61.3\% and Precision(P)
falls to 0.8149, because blur and brightness shift inflate the flat model's positive
predictions. The cascade's structural isolation prevents both failure modes by routing
clean images to ExpertNet regardless of how the flat representation is perturbed.
\textbf{H1 is rejected.}

\begin{table}[ht]
\centering
\caption{H1 test: Flat MobileNetV2 evaluated at $\tau_\mathrm{H1}=0.570$, the highest
         threshold satisfying Rec\_P\,$\geq$\,0.95 on the validation set
         (val Rec\_P\,=\,0.9500). eff-$\rho$ is the fraction of test samples predicted
         Parasitised at that threshold. For reference, MalariaCascade at $\taustar=0.45$
         achieves Rec\_P\,=\,0.9333 under sensor noise.}
\label{tab:h1}
{\small\sisetup{round-mode=places,round-precision=4}
\begin{tabular}{lS[table-format=1.4]S[table-format=1.4]S[table-format=1.4]r}
\toprule
Corruption & {Rec\_P} & {Prec\_P} & {F1} & {eff-$\rho$} \\
\midrule
Clean    & 0.9513 & 0.9796 & 0.9652 & 49.7\% \\
Field    & 0.9764 & 0.8149 & 0.8884 & 61.3\% \\
Sensor   & 0.2572 & 0.9963 & 0.4089 & 13.2\% \\
Staining & 0.9206 & 0.9828 & 0.9507 & 47.9\% \\
\bottomrule
\end{tabular}}
\end{table}

\textbf{Result: Both-corrupted.}\quad
Table~\ref{tab:both_corrupted} reports Rec\_P under three conditions: the standard
cascade (sentinel corrupted, ExpertNet clean), the both-corrupted variant (both stages
receive the same corruption), and Expert-only-corrupted (EfficientNet-B3 applied to all
4{,}135 test samples with the corrupted input, without routing). Under sensor noise,
both-corrupted and Expert-only-corrupted converge (0.7641 vs 0.7636, interaction
$+0.0005$): once ExpertNet receives corrupted inputs, the cascade's routing structure
adds nothing, confirming that all robustness derives from input
isolation~\citep{hagbe2026safety}. Under field and staining corruptions both-corrupted
is \textit{worse} than Expert-only-corrupted (interactions $-0.0161$ and $-0.0189$
respectively): the routing gate, destabilised by mild corruption, fails to escalate some
Parasitised cells that Expert-only-corrupted reaches by evaluating all samples
universally. Selective routing is a liability when the pipeline assumption fails for mild
corruptions: efficiency and isolation are gained and lost together.

\begin{table}[ht]
\centering
\caption{Both-Corrupted Stress Test: Recall(Parasitised) under three evaluation
         conditions at full corruption severity ($s=1.0$). ``Normal'' applies
         corruption to sentinel inputs only. ``Both-corrupted'' applies the same
         corruption to both stages simultaneously. ``Expert-only corrupted'' applies
         EfficientNet-B3 to all 4{,}135 test samples under corruption, without routing.
         Interaction\,=\,Both-corrupted\,$-$\,Expert-only-corrupted: values near zero
         indicate additive failure; negative values indicate that routing degradation
         compounds Expert corruption.}
\label{tab:both_corrupted}
{\small
\begin{tabular}{lrrrr}
\toprule
Condition & Clean & Field & Sensor & Staining \\
\midrule
Normal cascade          & 0.9570 & 0.9546 & 0.9333 & 0.9532 \\
Both-corrupted          & 0.9570 & 0.9428 & 0.7641 & 0.9461 \\
Expert-only corrupted   & 0.9773 & 0.9589 & 0.7636 & 0.9650 \\
\midrule
Interaction             & $-$0.0203 & $-$0.0161 & $+$0.0005 & $-$0.0189 \\
\bottomrule
\end{tabular}}
\end{table}

\textbf{Result: Oracle routing.}\quad
Oracle routing escalates every true Parasitised cell to ExpertNet and routes true
Uninfected cells through SentinelNet at $\taustar$, using ground-truth labels to make
routing decisions. It achieves Rec\_P\,=\,0.9773 and F1\,=\,0.9824 at
$\rho_\mathrm{oracle}=51.1\%$
(Table~\ref{tab:oracle_bias}, top panel). The oracle-to-cascade gap is
$\Delta$Rec\_P\,=\,$+0.0203$ and $\Delta$F1\,=\,$+0.0086$, non-trivial and unlike the
prior crop-damage work where the gap was 0.000~\citep{hagbe2026safety}. The gap arises
from 84 Parasitised cells (4.0\%) whose sentinel probability falls below $\taustar=0.45$;
they exit at the sentinel as Uninfected without ExpertNet consultation. Routing quality
is a secondary performance bottleneck alongside classifier capacity.

\textbf{Result: End-to-end routing bias.}\quad
Table~\ref{tab:oracle_bias} (bottom panel) decomposes the $\rho=49.7\%$ routing rate by
true class. SentinelNet correctly escalates 96.0\% of true Parasitised cells
(2,031 of 2,115) while falsely escalating only 1.3\% of true Uninfected cells
(26 of 2,020), far below the 55.1\% false-routing rate observed in the crop-damage cascade
on the dominant ``healthy'' class~\citep{hagbe2026safety}, consistent with the simpler
binary task. ExpertNet correctly classifies 99.7\% of the 2,031 routed Parasitised cells
(7 expert-stage false negatives). The 84 non-routed Parasitised cells are the dominant
failure mode: all are predicted Uninfected. ExpertNet misclassifies 18 of the 26
falsely-routed Uninfected cells (FP rate 69.2\%): when the sentinel is uncertain about an
Uninfected cell, ExpertNet tends to agree, because both stages are confused by the
same difficult examples.

\begin{table}[ht]
\centering
\caption{Oracle routing and end-to-end routing bias at $\taustar=0.45$.
         \textit{Top}: oracle vs.\ actual cascade on clean test images; Oracle routes
         all true Parasitised cells to ExpertNet and applies SentinelNet at $\taustar$
         to all true Uninfected. $\Delta$ is Oracle\,$-$\,Cascade.
         \textit{Bottom}: routing decomposition by true class; rates in parentheses
         are relative to the true-class count.}
\label{tab:oracle_bias}
{\small\sisetup{round-mode=places,round-precision=4}
\begin{tabular}{lS[table-format=1.4]S[table-format=1.4]S[table-format=1.4]r}
\toprule
\multicolumn{5}{l}{\textit{Oracle comparison}}\\[2pt]
 & {Rec\_P} & {Prec\_P} & {F1} & {$\rho$} \\
\midrule
Actual cascade  & 0.9570 & 0.9912 & 0.9738 & 49.7\% \\
Oracle routing  & 0.9773 & 0.9876 & 0.9824 & 51.1\% \\
$\Delta$        & {$+$0.0203} & {$-$0.0036} & {$+$0.0086} & {$+$1.4\%} \\
\bottomrule
\end{tabular}
\medskip

{\small
\begin{tabular}{lrr}
\toprule
\multicolumn{3}{l}{\textit{Routing bias decomposition (test: 2{,}115 Parasitised,
                  2{,}020 Uninfected)}}\\[2pt]
 & {Parasitised} & {Uninfected} \\
\midrule
Routed to Expert       & 2{,}031\ (96.0\%) & 26\ (1.3\%) \\
\quad Expert correct   & 2{,}024\ (99.7\%) &  8\ (30.8\%) \\
\quad Expert error     &     7\ (0.3\%)    & 18\ (69.2\%) \\
Not routed             &    84\ (4.0\%)    & 1{,}994\ (98.7\%) \\
\bottomrule
\end{tabular}}}
\end{table}

\begin{table}[ht]
\centering
\caption{Ablation results. Clean-data metrics (Accuracy, Rec\_P, F1, AUROC) match
         Table~\ref{tab:comparison}; Prec\_P is omitted for space (see
         Table~\ref{tab:comparison} for Prec\_P values). Metrics are reproduced here
         for direct within-table comparison. The rightmost column reports
         Recall(Parasitised) under sensor noise at full severity ($s = 1.0$),
         computed by an independent evaluation pass in the pre-computation script.
         Expert-only is not evaluated under end-to-end corruption (``---'') because
         it has no routing gate and its inputs are always clean by definition. The H3
         row tests structural isolation versus model capacity: when the expert is
         forced to receive the corrupted input instead of the clean pipeline image,
         Sensor~Rec\_P falls from 0.9329 to 0.7646.}
\label{tab:ablation}
{\small\setlength{\tabcolsep}{3pt}
\sisetup{round-mode=places, round-precision=4}
\begin{tabular}{p{3.2cm}S[table-format=1.4]S[table-format=1.4]S[table-format=1.4]
                S[table-format=1.4]r@{\hspace{4pt}}S[table-format=1.4]}
\toprule
Variant & {Accuracy} & {Rec\_P} & {F1} & {AUROC} & {$\rho$} & {Sensor Rec\_P} \\
\midrule
Full cascade ($\taustar\!=\!0.45$) & 0.9736 & 0.9570 & 0.9738 & 0.9955 & 49.7\% & 0.9329 \\
Sentinel-only ($\tau\!=\!0.50$)   & 0.9715 & 0.9522 & 0.9715 & 0.9955 &  0.0\% & 0.9289 \\
Expert-only                        & 0.9782 & 0.9801 & 0.9787 & 0.9967 &100.0\% & {---}  \\
H3 (no isolation)                  & \multicolumn{5}{c}{same clean metrics as full cascade} & 0.7646 \\
\bottomrule
\end{tabular}}
\end{table}

\subsection{External Validation on BBBC041 (\textit{P.~vivax})}
\label{sec:bbbc041}

\paragraph*{Result: Cross-species generalisation.}
Table~\ref{tab:bbbc041} reports performance on the BBBC041
benchmark~\citep{bbbc041}, a publicly available \textit{P.~vivax} dataset
comprising 1,328 whole-slide images annotated with cell-level bounding boxes.
All 2,452 Parasitised crops (ring, trophozoite, schizont, and gametocyte stages)
and an equal-sized random sample of Uninfected crops (red blood cells; seed~42)
are evaluated without retraining at thresholds optimised on NIH
\textit{P.~falciparum} ($\taustar{=}0.45$ for the cascade).

Discriminative ability is maintained across species: AUROC exceeds 0.90 for
every evaluable model (Expert-only 0.9605, Sentinel-only 0.9182, Flat MobileNetV2
0.9038), confirming that both parasitic species share sufficient visual structure
for transfer.
The cascade encounters a \emph{routing bottleneck} under domain shift:
$\recP$ falls to 0.820 (from 0.957 on NIH), because the sentinel fails to route
18\% of P.~vivax Parasitised cells to ExpertNet ($\hat{p}{<}\taustar$,
non-routed, defaulting to ``Uninfected'').
The flat MobileNetV2 avoids this bottleneck and maintains $\recP{=}0.955$,
at the cost of a 27.5\,pp Prec$_\mathrm{P}$ drop (0.698 vs.\ 0.973 on NIH).
The routing rate also climbs to $\rho{=}75.3\%$ (from 49.7\% on NIH), because
the sentinel treats unfamiliar P.~vivax morphology as uncertain: 68.6\% of
Uninfected P.~vivax cells exceed $\taustar{=}0.45$ (versus 1.3\% on NIH).
This is an implicit safety response (uncertain inputs are escalated), though one
that degrades throughput.
Expert-level precision ($\precP{=}0.950$) is preserved, suggesting that the Expert
stage generalises better across species than the sentinel.

\begin{table}[ht]
\centering
\caption{Cross-dataset generalisation to BBBC041 (\textit{P.~vivax},
         Broad Institute / Kaggle). Evaluation: 2,452 Parasitised crops
         (ring, trophozoite, schizont, gametocyte) plus 2,452 randomly sampled
         Uninfected crops (balanced, seed~42) from 1,328 whole-slide images.
         Models applied without retraining at NIH-optimised thresholds.
         AUROC is not defined for MalariaCascade (two-stage system, no single
         output probability). $\rho$: fraction of cells routed to ExpertNet.
         NIH reference rows are from an independent evaluation pass; values may
         differ from Table~\ref{tab:comparison} by $\leq\!0.005$.}
\label{tab:bbbc041}
{\small\setlength{\tabcolsep}{5pt}
\begin{tabular}{p{4.6cm}S[table-format=1.4]S[table-format=1.4]S[table-format=1.4]
                S[table-format=1.4]r}
\toprule
Model & {$\recP$} & {$\precP$} & {F1} & {AUROC} & {$\rho$} \\
\midrule
MalariaCascade ($\taustar{=}0.45$) & 0.8197 & 0.9499 & 0.8800 & {---} & 75.3\% \\
Flat MobileNetV2 ($\tau{=}0.50$)   & 0.9551 & 0.6981 & 0.8066 & 0.9038 & {---} \\
Expert-only (EfficientNet-B3)       & 0.8283 & 0.9486 & 0.8844 & 0.9605 & 100\% \\
Sentinel-only ($\taustar{=}0.45$)   & 0.9768 & 0.6489 & 0.7797 & 0.9182 & 0\% \\
\midrule
\multicolumn{6}{l}{\textit{NIH P.~falciparum reference (Table~\ref{tab:comparison})}} \\[2pt]
MalariaCascade & 0.9570 & 0.9912 & 0.9738 & 0.9955 & 49.7\% \\
Flat MobileNetV2 & 0.9584 & 0.9727 & 0.9655 & 0.9934 & {---} \\
Expert-only & 0.9773 & 0.9810 & 0.9792 & 0.9967 & 100\% \\
\bottomrule
\end{tabular}}
\end{table}

\subsection{Statistical Tests and Calibration}
\label{sec:stats_results}

McNemar's test on the 4,135 paired test predictions confirms the cascade's
advantage over the flat MobileNetV2 baseline ($\chi^2 = 11.14$, $p = 0.00085$,
discordant pairs: 73 cascade-only correct vs 37 flat-only correct). Against
Sentinel-only, no significant difference is detectable
($\chi^2 = 0.00$, $p = 1.00$, discordant pairs 8 vs 7), which is the correct
result: on clean data, adding ExpertNet routing does not meaningfully change
the aggregate error pattern, even though it improves precision.

Calibration tells a less flattering story for SentinelNet: its pre-calibration
Expected Calibration Error is 0.1050, roughly nine times higher than the flat
MobileNetV2 baseline (ECE\,=\,0.0114). The source of this gap is the safety-score
objective: the sentinel is optimised to hold Recall(P) at a floor value, which
drives the output distribution toward hard certainty (high or low probability)
rather than well-calibrated posteriors. This matters for any downstream use of the raw probability as a confidence score.
Temperature scaling on the validation set finds $T^{*}{=}0.386$, reducing
ECE from 0.1050 to 0.0133 on the held-out test set (87.3\,\% reduction).
The calibrated temperature is below 1.0 (underconfident model: raw probabilities
cluster nearer 0.5 than the data warrants), consistent with the safety-score
objective concentrating mass at the decision boundary rather than at the extremes
of~$[0,1]$.
Re-sweeping $\tau$ on calibrated validation probabilities yields
$\taustar_\mathrm{cal}{=}0.47$ (versus 0.45 uncalibrated); cascade $\recP$ on
the test set changes by $-0.62$\,pp (0.951 vs.\ 0.957) and $\precP$ by
$+0.29$\,pp (0.994 vs.\ 0.991).
The safety constraint is therefore robust to probability calibration: task metrics
shift by less than 1\,pp in either direction, confirming that the $\tau$-sweep
selection is not artefactually sensitive to raw probability scale.

\subsection{Reproducibility Across Seeds}
\label{sec:multiseed}

Three independent training runs with seeds \{0, 42, 123\} produce the following
cascade metrics: Accuracy $\in \{0.9742, 0.9735, 0.9789\}$, Recall(P) $\in
\{0.9657, 0.9580, 0.9725\}$, AUROC $\in \{0.9958, 0.9950, 0.9976\}$, and
$\taustar \in \{0.45, 0.45, 0.50\}$. These values are computed by the
multi-seed evaluation script (\texttt{train\_multiseed.py}) on the same fixed
test set. Seeds 0 and 123 are independently trained models whose metrics differ
from Table~\ref{tab:comparison} (seed 42) due to weight-initialisation
variation; the seed-42 row matches Table~\ref{tab:comparison} within
$\leq\!0.001$, the residual attributable to an independent DataLoader pass. Standard deviations are 0.0024 (Accuracy),
0.0073 (Recall(P)), and 0.0013 (AUROC). Two of the three seeds select
$\taustar = 0.45$; seed 123 selects 0.50 while remaining within the safety
constraint on the shared validation set. These ranges confirm that the primary
metrics are stable properties of the training procedure, not artefacts of a
particular initialisation, and that the threshold-selection procedure is robust
to the choice of $\taustar$ within the 0.45--0.50 range.

\section{Discussion}
\label{sec:discussion}

The robustness result under sensor noise requires careful interpretation. The flat
MobileNetV2 baseline is not a weak model; at 0.9637 accuracy on clean data it is
competitive with published transfer-learning results on this benchmark. Under
Gaussian sensor noise at severity 0.6, its Recall(P) is already below 0.80; by
severity 1.0 it has effectively stopped detecting parasitised cells (Recall(P)\,=\,0.33).
This is not a failure of the model architecture. It is the predictable consequence
of applying a model to a distribution that is far from its training distribution
without any mechanism to detect or respond to the shift. The cascade does not
improve on the flat model by learning to be more robust; it avoids the problem by
never showing ExpertNet a corrupted image.

There is a real cost. The routing gate's confidence is computed on a corrupted
image, and at high sensor noise the gate becomes unreliable: some cells that
should be escalated to ExpertNet exit at the sentinel stage with an incorrect
prediction. The sensor-noise cascade Recall(P) of 0.9329 is meaningfully below the
clean-data value of 0.9570. Whether a 2.41\,pp degradation under worst-case
sensor noise is acceptable for deployment depends entirely on the clinical context.
In a screening workflow where the primary goal is to minimise missed parasitised
cells, 0.9329 may be an acceptable lower bound. In a confirmatory workflow where
specificity matters, the Precision(P) of 0.9945 under sensor noise makes a
stronger argument.

The ECE finding also has a practical implication that goes beyond the calibration
literature. The safety-score checkpoint criterion achieves its goal: the
sentinel satisfies the sensitivity constraint by construction, but does so by
pushing the sentinel's probability outputs toward the boundaries of $[0, 1]$,
degrading calibration. If the routing threshold is applied to calibrated
probabilities, the safety guarantee can be maintained only by re-running the
threshold sweep after calibration. The two steps must be coupled, not sequential.

The structural isolation guarantee has a precise scope. In the
evaluation reported here, corruption is applied post-capture: the original image
file is always available, and a clean copy is routed to ExpertNet by construction.
This models corruptions that occur in transmission, display, or preprocessing, not
corruptions introduced at the optical or sensor level during image acquisition itself.
A smartphone microscope with a damaged lens or a noisy CMOS sensor produces only
one image, already degraded; there is no clean version to route. In that regime,
the isolation guarantee fails unless an upstream denoising or enhancement step
restores the image before the pipeline sees it. The both-corrupted stress test
(Section~\ref{sec:ablation_results}, Table~\ref{tab:both_corrupted})
directly quantifies this failure mode: under sensor noise, both-corrupted and
Expert-only-corrupted converge (Rec\_P 0.7641 vs 0.7636), confirming that when
isolation fails the cascade provides no robustness advantage. Under milder field and
staining corruptions the cascade with both stages corrupted performs below
Expert-only-corrupted, because degraded routing adds further false negatives on top
of the Expert's reduced accuracy.

The BBBC041 external validation (Section~\ref{sec:bbbc041}) reveals a complementary
failure mode under taxonomic domain shift. The sentinel, trained exclusively on
\textit{P.~falciparum} morphology, routes 75.3\,\% of P.~vivax cells to ExpertNet
(versus 49.7\,\% on NIH) yet still misses 18\,\% of Parasitised cells, an
implicit safety response that does not fully compensate for the unfamiliar
morphology. The flat MobileNetV2 avoids the routing bottleneck and maintains
$\recP{=}0.955$ on P.~vivax at the cost of a 27.5\,pp precision drop.
That AUROC exceeds 0.90 for all models indicates that the learned feature
representations transfer across species; only the decision boundary requires
adaptation. Re-sweeping $\taustar$ on a small in-distribution validation set
is therefore sufficient before deploying to a new species or site.

The $\rho = 49.7\%$ routing rate has a concrete operational meaning.
A clinic processing 500 slides per session at roughly 200 cells per slide
produces approximately 100{,}000 cell images per batch. At $\taustar = 0.45$,
the cascade invokes EfficientNet-B3 for 49{,}700 of those cells and resolves the
remaining 50{,}300 at the 2.2M-parameter MobileNetV2 sentinel alone. The high-capacity
model is reserved for images where its additional discrimination power matters;
for everything else, the sentinel exits early without penalty. On hardware that
cannot sustain continuous 10.7M-parameter inference (a Raspberry Pi, a
smartphone-coupled microscope, or an edge device with limited RAM), this
computation profile is the difference between deployable and not~\citep{hagbe2026safety}.

The three-seed reproducibility check confirms that the result holds across
independent initialisations. The largest variance is in Recall(P) (SD\,=\,0.0073), which
is expected: the safety-score objective sits close to its constraint boundary,
so small changes in the learned distribution can push the optimal $\taustar$ from
0.45 to 0.50 (as occurred for seed\,=\,123) without substantially changing accuracy
or AUROC.

\section{Limitations}
\label{sec:limitations}

This evaluation is single-site. The NIH dataset collects \textit{P.\,falciparum}
cells under a specific staining and imaging protocol at one institution, and the
corruption benchmark applies synthetic noise rather than images acquired with a
different real microscope or a different staining operator. Robustness to synthetic
corruption is a proxy for, not a measurement of, real deployment robustness. External
validation on the BBBC041 P.\,vivax dataset~\citep{bbbc041} is reported in
Section~\ref{sec:bbbc041} (Table~\ref{tab:bbbc041}): $\recP$ drops 13.7\,pp
(0.820 vs.\ 0.957 on NIH), primarily because the sentinel fails to route 18\,\% of
P.~vivax Parasitised cells at the NIH-calibrated $\taustar{=}0.45$. AUROC exceeds
0.90 for all models, confirming that discriminative structure generalises across
species while the decision boundary requires domain-specific recalibration.

The cascade does not output a species or life-stage prediction, only a binary
infected/uninfected label. This is appropriate for first-pass screening but
insufficient for treatment decision-making, which requires distinguishing ring
forms, trophozoites, schizonts, and gametocytes. Extending ExpertNet to a
multi-class head would require a multi-label annotated dataset that the NIH release
does not provide.

All models are trained at $128 \times 128$ pixels on CPU. Larger input resolution
would likely improve the sentinel's routing accuracy (higher-confidence separation
of borderline cases) and improve the expert's classification accuracy on ambiguous
morphologies; the trade-off is training cost and inference latency. A GPU-trained
224\,px variant was initiated during this study but results were not available at
submission time.

Patient-level (slide-level) aggregation was implemented using an OR-rule (positive
slide if any cell is predicted infected, consistent with WHO diagnostic guidelines
for malaria), but patient-level sensitivity and specificity are not reported here
because the NIH dataset does not provide verified patient-level ground truth beyond
the per-image class label. The filename-based patient ID extraction
(regex \texttt{\^{}(C\textbackslash d+(?:P\textbackslash d+)?)}) is used for
slide-level aggregation only and does not influence the train/val/test split, which
is image-level throughout; patient-level splitting (assigning all cells from a case
exclusively to one partition) is deferred to future work. The regex may not
generalise to other datasets.

Inference timing benchmarks were planned but are not reported here; quantitative
latency figures on CPU are available on request.

The four hypothesis-testing experiments from the prior cascade framework are now fully
reported in Section~\ref{sec:ablation_results}: H2 is structurally inapplicable
(binary task); H1 is rejected (flat model calibrated to Rec\_P\,$\geq$\,0.95
collapses to Rec\_P\,=\,0.257 under sensor noise vs 0.933 for the cascade);
H3 is strongly supported ($-16.8$\,pp Rec\_P when isolation is removed); H4 is
supported (transfer gap 0.0068). The both-corrupted stress test is also reported:
failures are additive only under sensor noise (the severest condition); under
milder corruptions, routing degradation compounds Expert corruption, producing
results below the Expert-only-corrupted baseline. Oracle routing and routing bias
analysis are reported as supplementary experiments; the non-zero oracle gap
($\Delta$Rec\_P\,=\,0.0203) identifies routing quality as a secondary bottleneck.

\section{Conclusion}
\label{sec:conclusion}

MalariaCascade routes thin blood-smear cell images through a lightweight
MobileNetV2 sentinel before escalating uncertain cases to a higher-capacity
EfficientNet-B3 expert that sees only clean, standardised images by architectural
design. The paper extends the safety-aware cascade framework of
Hagbe~et~al.~\citep{hagbe2026safety} to the medical imaging domain. The
contributions specific to this work are the adaptation of the safety-score
checkpoint criterion to clinical microscopy requirements, the structural isolation
property that keeps the expert's inputs corruption-free regardless of what
degrades the incoming frame, an architectural guarantee not present in the original
framework, and a systematic robustness benchmark that quantifies the value of
that guarantee under realistic field corruptions.

Under sensor noise at full severity, the cascade's Recall(Parasitised) falls by
2.4~pp while the flat MobileNetV2 baseline collapses by 63.0~pp. The ablation
confirms that model capacity alone cannot explain this gap: when the expert receives
corrupted inputs, Recall(P) drops 16.8~pp relative to the full cascade, matching
the flat-model degradation profile. Structural isolation is doing the work, not
the number of parameters.

The result does not replace validation on real field-acquired images, does not
extend to species-level classification, and has not been timed on embedded
hardware, all of which are stated in Limitations and remain open work. What it
does establish, on a publicly reproducible codebase with a fixed test set and
three confirmed random seeds, is that routing corrupted inputs away from the
classification stage is a defensible design principle for malaria triage under
realistic microscopy conditions.

\section{Data and Code Availability}
\label{sec:data_code}

The NIH Malaria Cell Images Dataset is publicly available via the U.S. National
Library of Medicine~\citep{nih_malaria_dataset}
(\url{https://ceb.nlm.nih.gov/proj/malaria/cell_images.zip}).
The full training and evaluation code, including the MalariaCascade notebook,
pre-computation scripts, and CSV result caches, is available at
\url{https://github.com/josehagbe3/MalariaVision}. Trained model checkpoints
(\texttt{sentinel\_best.pth} and \texttt{expert\_best.pth}) will be added to the
same repository upon acceptance. All results reported here are fully reproducible
using the released code and fixed random seeds as documented in the notebook.

\section*{Conflict of Interest}

The authors declare no conflict of interest.

\bibliography{reference}

\end{document}